\documentclass[letterpaper, 10pt, conference]{ieeeconf}      

\IEEEoverridecommandlockouts                              

\usepackage{graphicx} 
\usepackage{svg}
\usepackage{booktabs}
\usepackage{stfloats}
\usepackage{amsmath} 
\usepackage{amssymb}  
\usepackage{babel}
\usepackage{array}
\usepackage{url}
\newcolumntype{C}[1]{>{\centering\arraybackslash}p{#1}}

\usepackage{xcolor}

\usepackage[hidelinks]{hyperref}
\usepackage{fancyhdr}

\title{\LARGE \bf
CoopScenes: Multi-Scene Infrastructure and Vehicle Data for Advancing Collective Perception in Autonomous Driving
}

\author{Marcel Vosshans$^{1,2}$, Alexander Baumann$^{1}$, Matthias Drueppel$^{3}$, Omar Ait-Aider$^{2}$,\\ Youcef Mezouar$^{2}$, Thao Dang$^{1}$ and Markus Enzweiler$^{1}$
\thanks{$^{1}$The authors are with the Institute for Intelligent Systems which is part of the Faculty of Computer Science and Engineering, University of Applied Sciences Esslingen, Germany
        {\tt\small \{marcel.vosshans, alexander.baumann, thao.dang, markus.enzweiler\}@hs-esslingen.de}}%
\thanks{$^{2}$The authors are with the Institut Pascal ISPR (Images, Perception Systems and Robotics), Universite Clermont Auvergne INP / CNRS, France
        {\tt\small \{youcef.mezouar, omar.ait-aider\}@uca.fr}}%
\thanks{$^{3}$The author is with the Center for Artificial Intelligence, Baden Württemberg Cooperative State University Stuttgart, Germany
        {\tt\small \{matthias.drueppel\}@dhbw-stuttgart.de}}%
}

\begin{document}
\bstctlcite{IEEEexample:BSTcontrol}
\maketitle
\thispagestyle{fancy}  
\fancyhf{}             
\renewcommand{\headrulewidth}{0pt}
\renewcommand{\footrulewidth}{0pt}
\fancyfoot[C]{\footnotesize
  \begin{minipage}{0.95\textwidth}
    \centering
    \textcopyright~2025 IEEE.
    Personal use of this material is permitted. Permission from IEEE must be obtained for all other uses, in any current or future media, including reprinting/republishing this material for advertising or promotional purposes, creating new collective works, for resale or redistribution to servers or lists, or reuse of any copyrighted component of this work in other works.\\
    Accepted to be published at the 2025 IEEE Intelligent Vehicles Symposium (IV), Cluj-Napoca, Romania, June 2025.
  \end{minipage}
}
\thispagestyle{fancy}
\pagestyle{empty}

\begin{abstract}
The increasing complexity of urban environments has underscored the potential of effective collective perception systems.
To address these challenges, we present the CoopScenes dataset, a large-scale, multi-scene dataset that provides synchronized sensor data from both the ego-vehicle and the supporting infrastructure.

The dataset provides \(\mathbf{104}\) minutes of spatially and temporally synchronized data at \(\mathbf{10}\) Hz, resulting in 62,000 frames. 
It achieves competitive synchronization with a mean deviation of only \(\mathbf{2.3}\) ms. 
Additionally the dataset includes a novel procedure for precise registration of point cloud data from the ego-vehicle and infrastructure sensors, automated annotation pipelines, and an open-source anonymization pipeline for faces and license plates.
Covering nine diverse scenes with \(\mathbf{100}\) maneuvers, the dataset features scenarios such as public transport hubs, city construction sites, and high-speed rural roads across three cities in the Stuttgart region, Germany. 
The full dataset amounts to \(\mathbf{527}\) GB of data and is provided in the \textit{.4mse} format, making it easily accessible through our comprehensive development kit.

By providing precise, large-scale data, CoopScenes facilitates research in collective perception, real-time sensor registration, and cooperative intelligent systems for urban mobility, including machine learning-based approaches.
\end{abstract}

\section{Introduction}

Recent advancements in autonomous driving have primarily focused on the ego-vehicle, with comparatively less attention given to the interoperability of multiple intelligent agents. 
While datasets like KITTI \cite{geiger_vision_2013} or nuScenes \cite{caesar_nuscenes_2020} have significantly accelerated progress in ego-vehicle perception, establishing object detection, semantic segmentation, and scene completion as standard benchmarks.
Scaling autonomous and intelligent vehicle systems introduces novel opportunities, such as the potential to reduce the sensor setup required for individual vehicles by redistributing sensor capabilities to the infrastructure. 
This approach is based on the premise that highly accurate and comprehensive sensor setups are only necessary for complex scenarios, while most autonomous driving tasks can be executed using a simplified sensor configuration. 
By enabling shared access to infrastructure-based sensors, this paradigm offers several benefits: economic resource savings, accelerated market penetration of intelligent vehicles due to reduced purchase costs, and enhanced safety through fault-tolerant systems, such as the "two-out-of-three" redundancy principle.
However, such advancements also come with significant challenges. 
Effective multi-agent systems require addressing issues such as standardization, communication protocols, and the ability to contextualize and integrate observations across multiple agents.
Success in this domain depends on reliable algorithms for real-time spatial registration and robust infrastructure setups.

To address these challenges, we present a novel dataset capturing multiple scenes and maneuvers from two complementary perspectives: an intelligent vehicle and an observing infrastructure. 
The dataset ensures precise temporal and spatial synchronization, aligning both perspectives in a unified coordinate system.

\begin{figure}
    \centering
    \includegraphics[width=\linewidth]{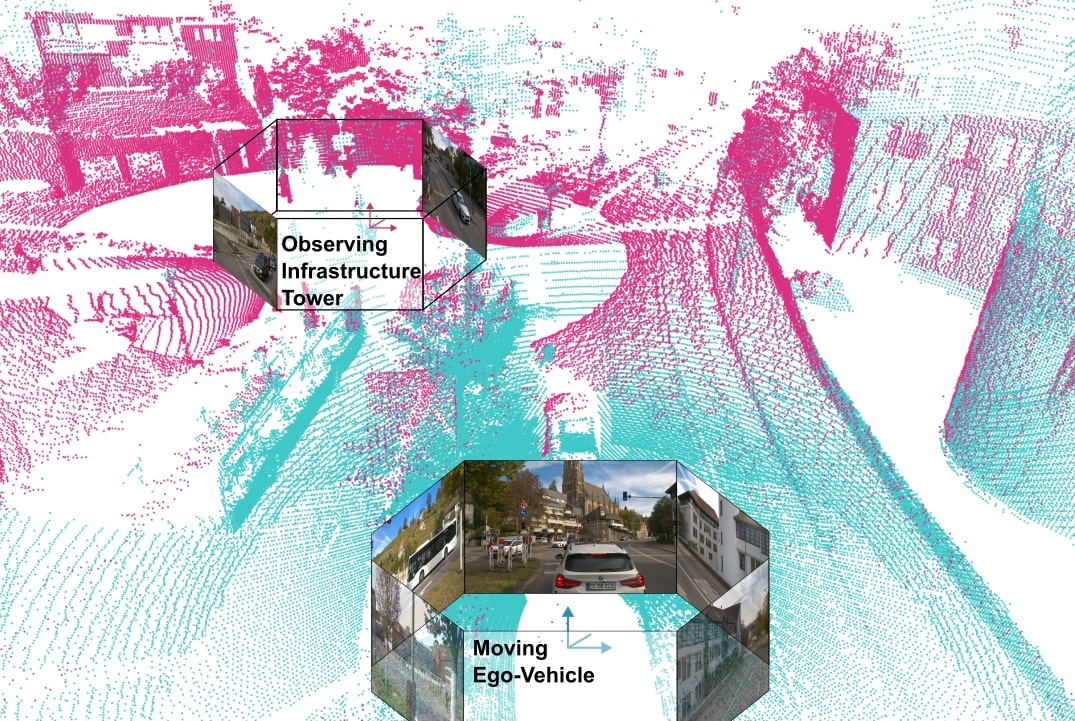}
    \caption{Presented is an extract from \textbf{CoopScenes, Scene \(\mathbf{4}\)}. The camera and LiDAR sensor setup of both agents—the observing infrastructure tower (pink) and the moving ego-vehicle (turquoise)—is transformed into a global coordinate system, with their respective camera perspectives overlaid.}
    \label{fig:3d_rgb_pointcloud}
\end{figure}

This dataset is tailored for public transportation scenarios, assuming fixed-route bus systems as manageable domains for equipping infrastructure.
It introduces novel research tasks, including collective perception, fault-tolerant decision-making, and real-time spatial registration, with a particular focus on the latter. 
By fostering the development of algorithms aimed at enhancing multi-agent interoperability and infrastructure-supported perception, this dataset lays the groundwork for advancing autonomous driving beyond ego-vehicle-centric paradigms.

\subsection{Contributions}
\begin{enumerate}
    \item \textbf{Multi-Scene Dataset} of synchronized vehicle and infrastructure perception data:
    \begin{itemize}
        \item \textbf{\(104\) minutes of data:} Collected across three cities in the Stuttgart region (Germany), featuring \(9\) scenes and \(100\) maneuvers.
        \item \textbf{Comprehensive sensor setup:} Includes ego-bus and infrastructure tower equipped with \(8\) cameras, \(6\) LiDARs, GNSS, and motion sensors, temporal and spatially synchronized.
        \item \textbf{Annotations:} Anonymized data with comprehensive calibrations, frame-wise transformation matrices, and \(40k\) labels.
    \end{itemize}
    \item Open-source \textbf{Anonymization Model}: Designed to blur faces and license plates effectively.
\end{enumerate}

\subsection{Public Resources}
Alongside this paper, we also publish:
\begin{itemize}
    \item \textbf{CoopScenes Dataset}\footnote{\url{https://coopscenes.github.io}} + Pipelines, Validation labels 
    \item \textbf{Development Kit}\footnote{\url{https://pypi.org/project/CoopScenes}} Library
    \item \textbf{BlurScene Model}\footnote{\url{https://github.com/CoopScenes/BlurScene}} Repository
\end{itemize}

\subsection{Paper Organization}
First, we introduce the dataset with a comprehensive overview, including detailed statistical measures and insights into its composition. Next, we describe the annotation algorithms employed and the techniques used to ensure the anonymization of personal data. Finally, we outline the range of tasks and applications for which the dataset is designed, highlighting its potential for advancing research in autonomous systems and multi-agent interoperability.
\section{Related Work}
\begin{table*}[ht!]
\centering
\caption{Comparison of vehicle and infrastructure-synchronized datasets based on selected features. *For some datasets, certain key figures were challenging to identify due to the unavailability of public access to the data.}
\begin{tabular}{p{0.145\textwidth}cccccc}
\textbf{Dataset} & \textbf{V2X-Sim} & \textbf{DAIR-V2X-C} & \textbf{V2X-Seq (SPD)} & \textbf{HoloVIC} & \textbf{TUMTraf V2X} & \textbf{CoopScenes} (ours) \\
\toprule

\textit{Year} & 2022  & 2022 & 2023 & 2024 & 2024 & 2025 \\

\textit{Real World} & Simulation & Yes & Yes & Yes & Yes & Yes \\

\textit{Open Access} & Yes & No & No & No & Yes & Yes \\

\textit{Development Kit} & Yes & Yes & Yes & * & Yes & Yes \\

\textit{Global Localisation} & Known  & GPS/IMU + SLAM & GPS/IMU + SLAM & RTK + E-Compass & Scan-Matching & Scan-Matching \\

\textit{Number Sensors (I/V)} & 5/7 & 2/3 & 2/3 & \textbf{18/5} & 5/4 & \textbf{6/11} \\

\textit{Points per frame} & 25k & 232k & 232k & 430k & 189k & \textbf{565k} \\

\textit{Time Sync. [ms]} & \textbf{0} & async & * & * & 24.9 & 2.3 \\

\textit{Length [min] (frames)} & 33:18 (10k) & 30:00 (39k) & 25:12 (15k) & 80:00 (48k) & 1:20 (0,8k) & \textbf{104:06 (62k)} \\

\textit{Unique Positions} & 6  & * & * & 5 & 1 & \textbf{9} \\
\textit{Traffic Scenarios} & 2  & 1 & 1 & 2 & 1 & \textbf{7} \\
\textit{Traffic Violations} & No  & No & No & No & \textbf{Yes} & \textbf{Yes} \\


\textit{Scaled for AI} & No  & No & No & \textbf{ Yes} & No & \textbf{Yes} \\

\textit{2D/3D Boxes} & 233k  & 464k & 11k & \textbf{2,66M} & 30K & 40k$^{1}$(5,54M)$^{2}$ \\

\bottomrule
\end{tabular} \\
\label{tab:dataset_comparison}
\vspace{0.5em}
\footnotesize
\raggedright
\hspace*{2em} $^1$2D labels
\hspace*{2em} $^2$automatic generated 2D labels
\end{table*}

\subsection{Traffic Datasets}
To support interoperable datasets for V2V (Vehicle-to-Vehicle) or V2X (Vehicle-to-Everything) perception tasks, we categorized datasets based on the interaction complexity of agents into three distinct groups. 

The first group, ego-perception, involves a vehicle relying solely on its own data to perform local perception, as exemplified by the Waymo dataset \cite{sun_scalability_2020}.

The second group focuses on the interaction between one static and one moving agent. 
This includes datasets such as V2X-Sim \cite{li_v2x-sim_2022}, which provides simulated data from intersections across various areas with sensors for both vehicles and infrastructure. 
Similarly, the DAIR-V2X \cite{yu_dair-v2x_2022} dataset and its follow-up release, V2X-Seq (SPD) \cite{yu_v2x-seq_2023}, offer real-world data but are limited to a single scene. 
The HoloVIC \cite{ma_holovic_2024} dataset introduces large-scale data with e.g. trajectories and multi-view overlapping, and the TUMTraf V2X \cite{zimmer_tumtraf_2024} dataset, the first European dataset in this category, introduces traffic violations, although it is also constrained to a single intersection.
Our dataset additionally features a mobile infrastructure platform, enabling the recording of diverse traffic positions and scenarios.
We summarized the features of these datasets in Table \ref{tab:dataset_comparison}, offering a comprehensive comparison of their capabilities and use cases.

The final group encompasses interactions between two or more moving agents, representing the most challenging scenario for collective perception.
An example of a dataset in this category is V2V4Real \cite{xu_v2v4real_2023}, which provides temporally synchronized frames collected in the USA.

\subsection{Collective Perception}
To enable the interoperability of traffic agents, standards such as ETSI ITS-G5 propose the use of Collective Perception Messages (CPM) \cite{noauthor_intelligent_2023}. 
These messages transmit the agent's position (longitude, latitude, height, and optionally dynamics data) alongside detected objects to facilitate high-level (late fusion) collective perception.

Beyond this communication framework, sensor data can also be fused on a lower level (early fusion) through mathematical transformation \cite{tischler_enhanced_2005}.
For an individual agent, this entails transforming collected sensor data into a common origin using known extrinsics, thereby enabling data fusion. 
Neural networks can also achieve this; for instance, SpatialDETR \cite{doll_spatialdetr_2022} performs amodal 3D bounding box detection from multiple images using locally constant spatial encoding. 
Conversely, RelMobNet \cite{rajendran_relmobnet_2022} employs a Siamese network to estimate the camera pose of stereo image pairs based on image features.

However, with two moving agents, the challenge arises as the extrinsics between sensors constantly change. This necessitates a preceding step to determine their relative sensor poses. 
CoopDet3D \cite{zimmer_tumtraf_2024} addresses this by first fusing local sensors at a BEV/Voxel grid level and subsequently integrating infrastructure and vehicle data using this feature map to perform collective perception. 
These approaches rely on local, on-the-fly data fusion.

In environments with pre-existing HD maps, all agents can localize themselves within a global reference frame and apply mathematical transformations for sensor fusion, provided the site is mapped.
For instance, VI-Map \cite{he_vi-map_2023} proposes a method for maintaining HD maps using road infrastructure observations, which supports local perception with global information via HD map localization.
\section{CoopScenes}
The concept behind the CoopScenes (Cooperative Scenes) is straightforward: how can the safety and reliability of autonomous vehicles be improved without relying exclusively on advancements in ego-vehicle technology?
The proposed solution is to enhance vehicle capabilities through infrastructure-based support systems.

For that, we focus on the public transportation system, as it provides a viable business case. 
Infrastructure sensors can be both technically and legally maintained by bus line operators. 
Additionally, buses typically operate within fixed and limited domains, making infrastructure planning more feasible.
As a result, we present a dataset featuring a movable infrastructure tower and a perception-equipped bus, collected across various public transport-related locations.

\begin{figure}
    \centering
    \includegraphics[width=\linewidth]{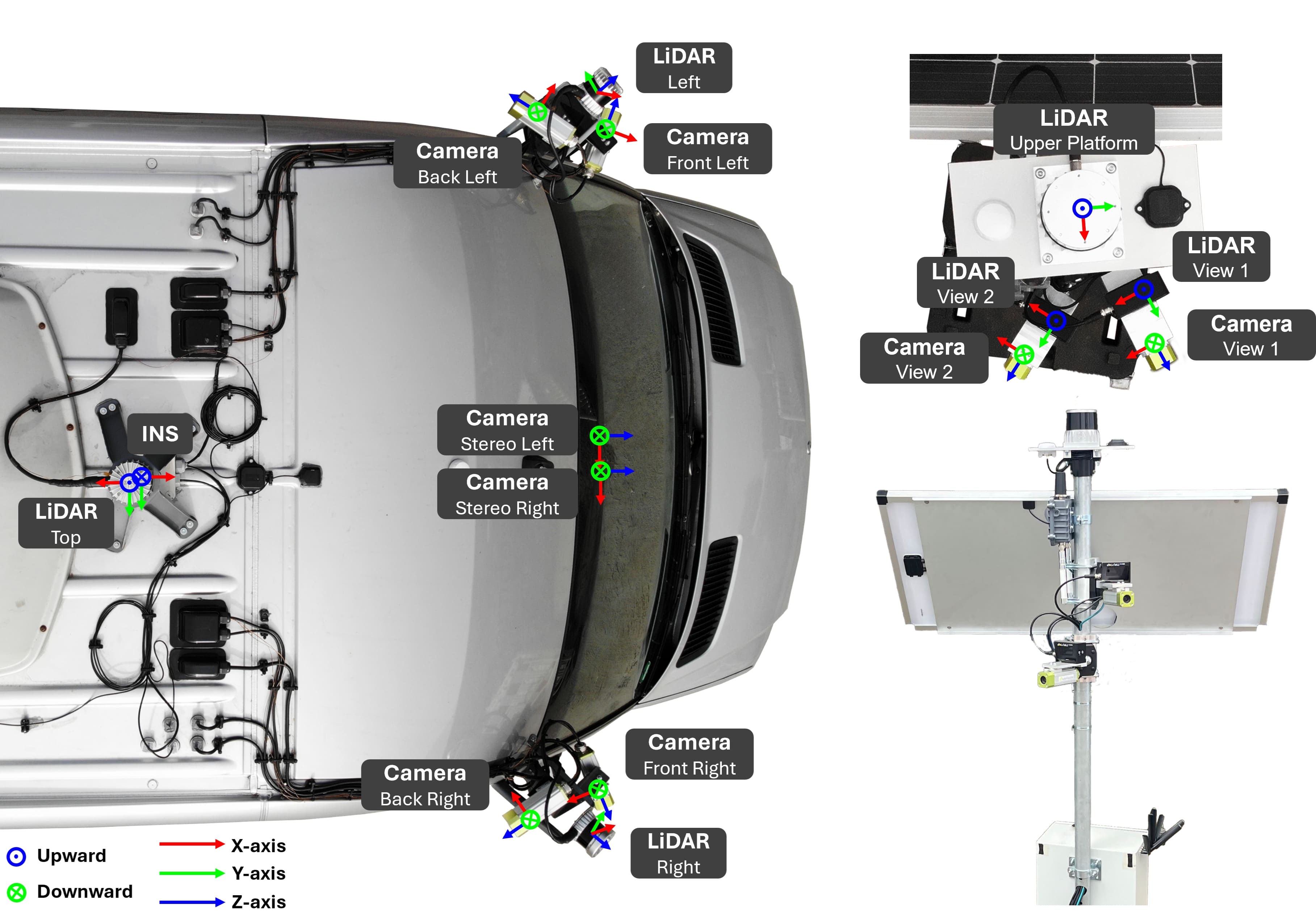}
    \caption{\textbf{Sensor setup} for the ego-vehicle and sensor tower, illustrated from a bird's-eye perspective with corresponding coordinate systems. Labels correspond to sensor types and individual instance names.}
    \label{fig:setup}
\end{figure}

\subsection{Agent Specification}
For this dataset, a small \textbf{public transport bus} (7 seats), as shown in Figure \ref{fig:setup}, was equipped with a comprehensive perception system consisting of:
\begin{itemize}
    \item Cameras: Four monocular cameras and one stereo pair (Basler a2A1920-51gmPRO, Full-HD).
    \item LiDAR: Three Ouster LiDARs—one OS-1 (\(128\) beams) mounted on top, and two OS-0 (\(128\) beams) inclined to cover areas close to the vehicle.
    \item Additional Sensors: GNSS, IMU, and vehicle data streams.
\end{itemize}

Similarly, the \textbf{infrastructure tower} was equipped with the following components, illustrated in Figure \ref{fig:setup}:
\begin{itemize}
    \item \(360\)° LiDAR: Ouster OS-2 (\(128\) beams) mounted on top.
    \item Two Flexible Arms: Each arm equipped with a Basler camera and a Blickfeld Cube1 solid-state LiDAR.
\end{itemize}

Both systems are temporally synchronized within a few milliseconds and operate at \(10\) Hz for all sensors, except the IMU and GNSS (\(1\) kHz).
All sensors are fully calibrated, both intrinsically and extrinsically, relative to their respective system origins. 

\textbf{Temporal Synchronization.}  
Synchronization is a critical factor in multi-agent sensor fusion at any level, and we have devoted special attention to ensuring precise temporal alignment.
Each agent in our dataset is equipped with a Precision Time Protocol (PTP) server, and all sensors within an agent are synchronized to this clock.
The PTP servers themselves are synchronized to the global GNSS time, achieving nanosecond (\(10^{-9}\) s) precision.

\begin{figure}
    \centering
    \includegraphics[width=\linewidth]{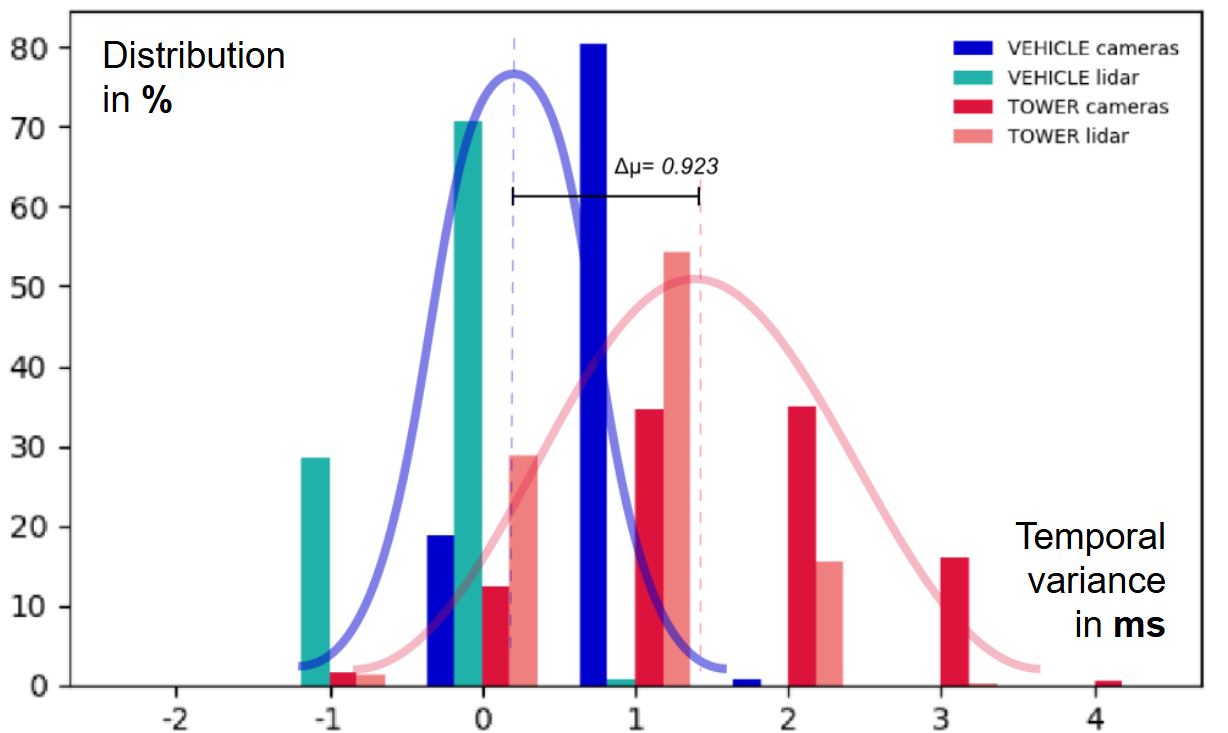}
    \caption{Using the vehicle's Top LiDAR as the reference, the capture \textbf{time differences of all sensors} across multiple scenes (\(4\)–\(7\)) are presented. The data is categorized into cameras and LiDAR, and color-coded by agent. On average, the infrastructure agent exhibits a delay of \(0.92\) ms, while the overall mean deviation is \(1.243\) ms. Normal distributions are included to illustrate the time delays for each agent.}
    \label{fig:time_sync}
\end{figure}

While using a common time source is essential, timing coordination is equally important. 
For our dataset, the TOP LiDAR of the vehicle serves as the primary trigger for all other sensors.
Vehicle sensors are triggered locally, while for the infrastructure (the second agent), we send an absolute timestamp to the tower. 
The tower sensors are then triggered at precise intervals (µs precision) of this timestamp every \(100\) ms. 
This setup results in an average timestamp difference of \(2.345\) ms between the TOP LiDAR and the other sensors.
A detailed analysis of the timestamp distribution is illustrated in Figure \ref{fig:time_sync}.

However, identical timestamps do not always correspond to identical real-world observations due to variations in sensor behavior (e.g. single image capture vs. continuous LIDAR measurements). 
To maximize the vehicle's synchronization, we rotated the TOP LiDAR by \(180\)° to ensure that the start of each LiDAR frame (trigger signal) aligns as closely as possible with the vehicle's front camera's image capture.

\begin{figure*}[t]
    \centering
    \includegraphics[width=0.98\textwidth]{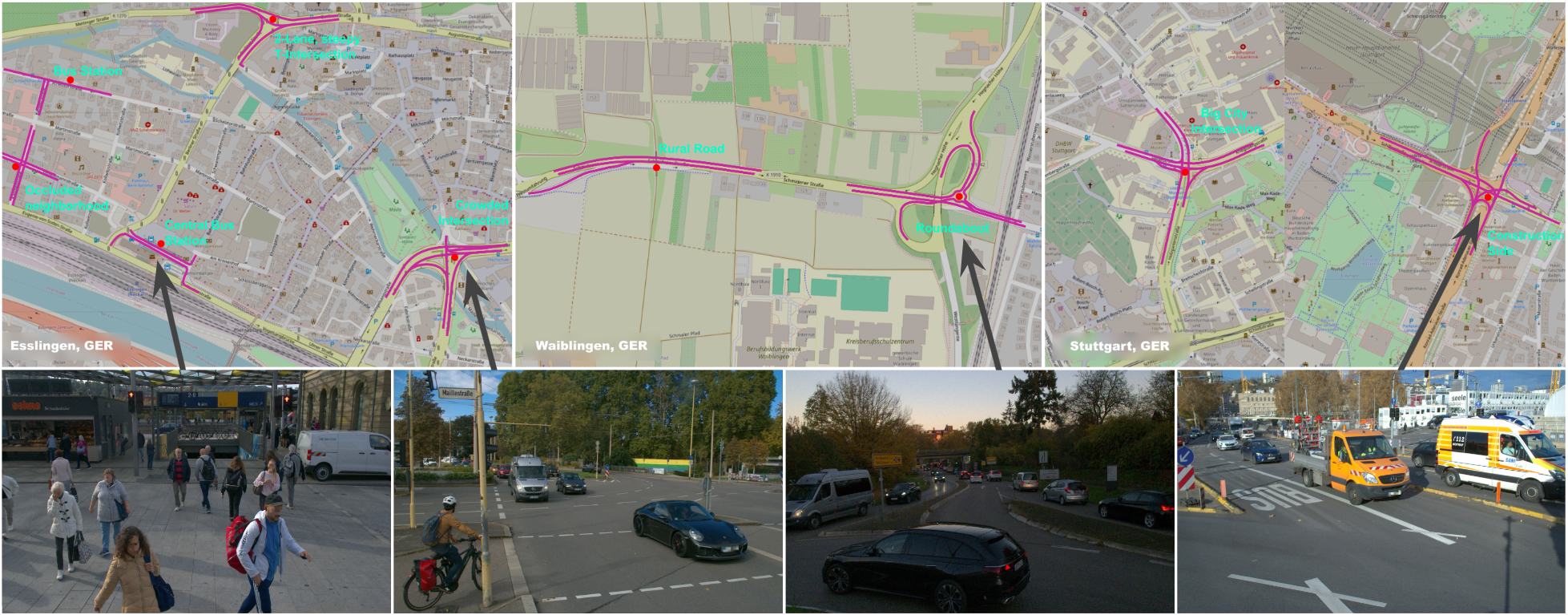}
    \caption{The upper row displays the locations where the scenes were captured. From left to right: Esslingen, Waiblingen, Stuttgart (DHBW/ Central Sation). The bottom row presents impressions from the scenes themselves, captured by the infrastructure camera. From left to right: the central bus station in Esslingen (Scene, \(5\)), a crowded intersection in Esslingen (\(1\)), a roundabout in Waiblingen during late dawn (\(7\)), and a construction site in the center of Stuttgart (\(9\)).}
    \label{fig:scene_book}
\end{figure*}

To ensure seamless interaction with out dataset, we provide an open-source Python development kit, precompiled and available on PyPI, offering a comprehensive research environment with a data loader and various assistive tools. 
The dataset is distributed in the \textit{.4mse} binary format, which can be easily accessed using the provided library. 
This enables intuitive data handling with just a few lines of code, allowing researchers to focus on their work rather than on data preparation.
Additionally, the toolkit includes assistive tools for visualization, sensor transformation, LiDAR motion compensation, hidden point removal, or LiDAR-to-camera projection.

\subsection{Dataset Collection}
For the dataset, we visited nine locations, which can be grouped into three distinct types of urban environments in the Stuttgart region, Germany, ranging from small villages to near-metropolitan areas.
In each location, multiple scenes were recorded, and within each scene, various maneuvers were captured multiple times.
Each recording required setting up the infrastructure tower, performing measurements, and conducting individual calibrations, including GNSS corrections and extrinsic calibration of the tower. 
This was necessary as the flexible sensor arms were adapted for optimal scene coverage at each location.
The locations covered in our scene book are visualized in Figure \ref{fig:scene_book}.
In total, the dataset includes \(104\) minutes of synchronized data from all sensors. 
The data is organized into \(9\) scenes with different \(100\) maneuvers,\ resulting in \(62,465\) frames, and a total data volume of \(527\) GB.
To manage the dataset size, we compressed the image data into \textit{.jpg} format and applied lossless compression to the LiDAR data by discarding missed measurements. 
The data is provided in \(674\) files, each approximately \(0.8\) GB in size, using our \textit{.4mse} format, which is easily accessible through our research and development kit.
The recorded scenes encompass public transport-related areas, including central bus stations, crowded urban zones, rural roads, roundabouts, densely occluded neighborhoods, and interactions with emergency and garbage collection vehicles.
Our recordings span the summer, fall, and winter seasons, with scenes captured in various lighting conditions, including early morning twilight and evening dusk.  However, no recordings were conducted during nighttime or under adverse weather conditions such as rain.
Some impressions of these scenarios are illustrated in Figure \ref{fig:scene_book}.

We provide a train/validation/test split for training neural networks, with an approximate ratio of \(75 / 15 / 10 \%\). 
To ensure the validity of learned tasks, we believe that it is important to avoid mixing scenes across the splits. 
This approach introduces entirely new scenarios in completely different environments to challenge the model's behavior across each split.
For this reason, the \textit{crowded intersection} in Esslingen serves as the validation scene, while the \textit{big city intersection} in Stuttgart is designated as the test scene.

\subsection{Purpose \& Tasks}
The dataset, like other ego-vehicle datasets, is suitable for tasks such as object detection, segmentation, or even SLAM. 
However, the uniqueness of our data lies in the availability of highly precise information from both the moving ego-vehicle and an observing infrastructure, which are spatially and temporally synchronized.
Another distinctive feature is the dataset's large scale and variety, making it particularly well-suited for applications involving neural networks.
With our dataset we aim to foster research in the following areas: 

\textbf{Collective Perception Tasks.}
With interoperability between traffic agents enabled, it becomes possible to enhance a vehicle's perception by detecting occluded objects or extending its field of view. 
This can be achieved using models like DiscoNet \cite{li_learning_2021} or CoopDet3D \cite{zimmer_tumtraf_2024} for object detection or CoBEVT \cite{xu_cobevt_2022} for cooperative semantic segmentation, which have so far been primarily trained on simulated or limited datasets.
Additionally, this approach allows for the validation of detections or planning decisions from alternative perspectives. 
Depending on the implementation, it can also serve as a fallback mechanism in the event of system failures.

\textbf{Sensor Fusion and Transformation.}
Since the transformation data in our dataset is calculated offline and incorporates all maneuver data as a unified information source, the dataset provides transformation matrices even in scenarios without sensor overlap. 
This enables investigations into real-time sensor registration confidence and the determination of working requirements or effective range. 
Inspired by RelMobNet \cite{rajendran_relmobnet_2022}, which predicts sensor transformations based on image features, the dataset also opens possibilities for exploring neural network-aided monocular registration.

\textbf{Sensor System Design Research.}
As introduced earlier, we posed the question: How minimal can the onboard vehicle sensor setup be while relying on infrastructural support in dense urban areas? 
To explore this, we equipped the vehicle with a comprehensive array of sensors. 
Using a functioning collective perception system, it becomes possible to incrementally reduce the onboard sensor setup and compare results, such as detection performance. 
This approach enables research on outsourcing sensors to the infrastructure.

\textbf{Behavioral Analysis and Prediction.}
From a self-supervised learning perspective, an intersection provides the opportunity to consistently observe the same location, enabling the system to learn local behaviors, such as an illegal U-turn — an action that a general autonomous vehicle might not anticipate.
Such scenarios can be effectively addressed with the assistance of local sensor infrastructure.
\section{Reference Label \& Ground Truth}
Scalability is a critical factor for this dataset, which is why we focus on scalable solutions for annotations and data preprocessing, such as anonymization. 
The aim is to maintain flexibility while ensuring the dataset remains easy to extend. 
Unlike a traditional bounding box detection benchmark, the primary focus of this dataset is on interoperability between infrastructure and intelligent vehicles.
The challenges may differ from those of current ego-vehicle datasets, and new challenges are likely to emerge, such as determining the transformation between agents in real time with limited data.

\subsection{Labeling Pipelines.}
Offline annotations offer significant advantages over real-time models, primarily due to the absence of time constraints for computation. 
This flexibility allows the application of advanced foundation models to enhance annotation precision. 
In object instance labeling, it becomes possible to reference frames before or after the target frame, ensuring consistent annotations or correcting partially occluded objects as demonstrated by Motional in their nuPlan dataset \cite{caesar_nuplan_2022}.
Since automatic labeling pipelines cannot guarantee ground truth, we refer to them as reference labels, signifying alignment with state-of-the-art precision.
Through automatic annotations, we introduce a learning cascade, recognizing potential discrepancies compared to the labels on which these models were trained. 
This principle extends equally to training processes based on the reference labels.
Despite these challenges, this approach presents opportunities to develop specialized models capable of achieving performance comparable to their reference foundation models while requiring less data or resources.

\begin{figure}[b]
    \centering    \includegraphics[width=\linewidth]{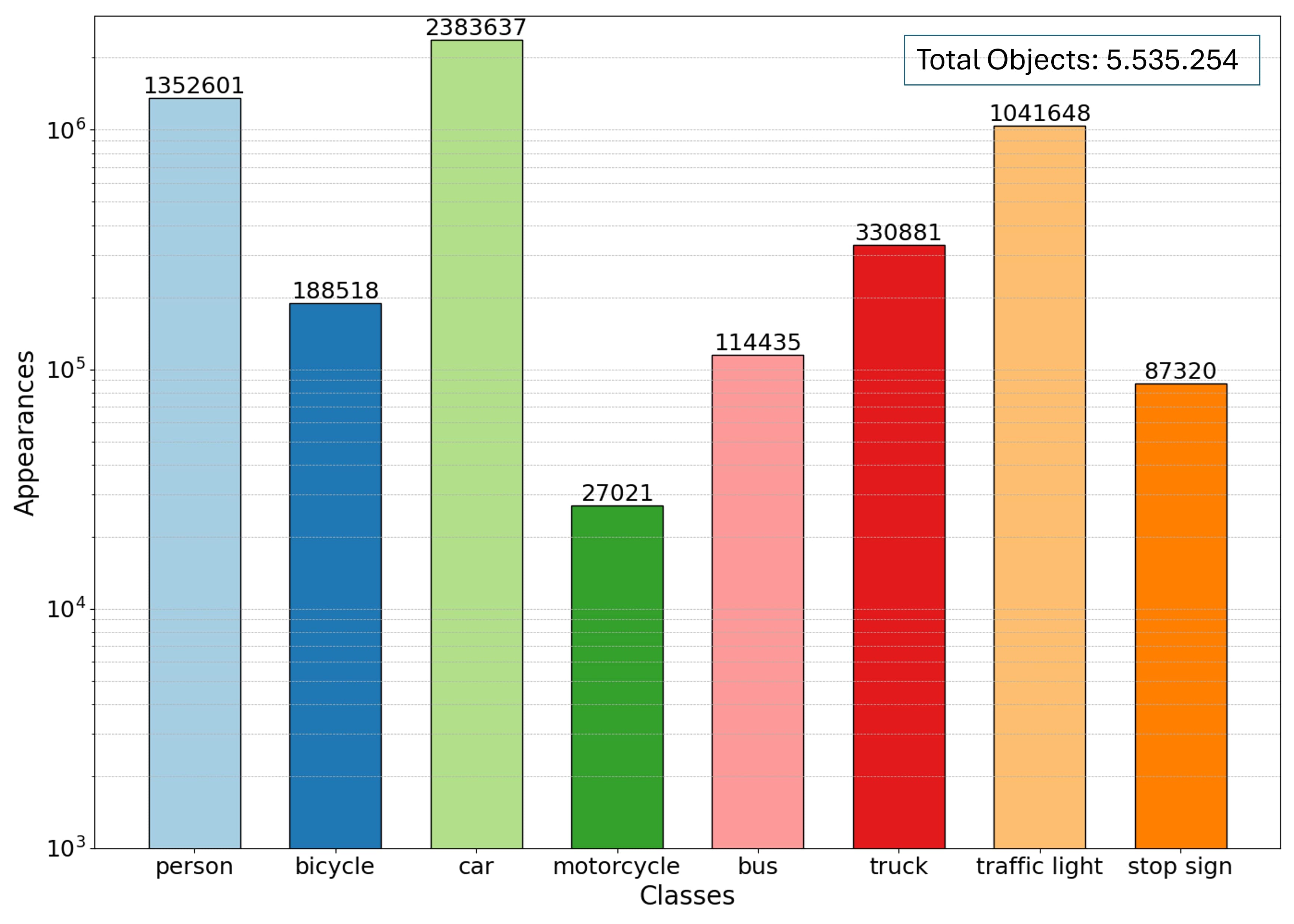}
    \caption{The \textbf{dataset analysis} highlights the distribution of object classes based on their occurrences. Instances were counted across all cameras and all frames, resulting in an average of \(12\) objects per frame per camera.}
    \label{fig:analysis}
\end{figure}

In this initial publication, we provide 2D bounding boxes, which were used to analyze the dataset.
We utilized a pretrained Co-DETR model \cite{zong_detrs_2023}, specifically the CO-DINO, ViT-L, COCO (instance) network, to detect potential objects and count classes such as persons, cars, traffic lights, and others.
Following the COCO mapping \cite{lin_microsoft_2014}, we included the classes depicted in Figure \ref{fig:analysis} as possible instances. 
To evaluate the precision of our annotation pipeline, we manually labeled a subset of the data and quantitatively assessed its precision. 
This precision comparison, along with the results, serves as our quantitative analysis of the dataset.
As shown in Figure \ref{fig:analysis}, a total of \(5.54\) million instances were detected, with an average of \(12.3\) objects per frame per camera, making all scenes quite crowded. 
The detected objects mainly consist of cars (\(42.9\%\)), persons (\(24.3\%\)), and traffic lights (\(18.8\%\)). 
Less frequent categories include trucks (\(6\%\)), bicycles (\(3.4\%\)), and the remaining classes—motorcycles, buses, and stop signs—together accounting for \(4.2\%\).
We will also publish the hand-annotated labels alongside annotation pipelines, such as those for 3D bounding boxes, to facilitate benchmarking of different models and evaluate potential improvements.
Our current results, achieved with Co-DINO, show a precision of \(90.6 \%\) and a recall of \(89.2 \%\) resulting in an F1-Score of \(89.4 \%\). These results were calculated for vehicles, which constitute the majority of detected objects in the dataset (\(51.1 \%\)), while excluding penalties for objects smaller than \(40\) pixels.

\subsection{Agent Extrinsics.}
One of the fundamental requirements for sensor interoperation is the transformation into a common coordinate system. 
This requires the translation \(t\) and rotation \(R\) of one sensor relative to another. 
In our approach, we defined the top sensor of the vehicle and the tower as the agent origins and calibrated all sensors of an agent extrinsically relative to this point. 
Consequently, only a single transformation is needed between the two agents, with the tower designated as the global origin \([0,0,0]\).
While GNSS provides an intuitive starting point for such transformations, it only offers a translation vector.
Rotation remains problematic, and GNSS precision can vary significantly in urban environments. 
Our approach addresses the transformation problem in a relative rather than global manner, simplifying the process and enabling derived transformation matrices to be transferred between frames, provided consistency is maintained.
We begin by employing the KISS-ICP algorithm \cite{vizzo_kiss-icp_2023}, which performs LiDAR odometry using frame-wise Iterative Closest Point (ICP) registration \cite{besl_method_1992} across all maneuvers.
Since the smallest GNSS distance between the vehicle and the tower likely corresponds to the greatest overlap in LiDAR point clouds, we first identify a batch of \(100\) frames with the shortest GNSS distances. For these frames, RANSAC \cite{fischler_random_1987} is used to obtain an initial transformation guess, which is then refined using ICP. 
The translation vectors of the estimated tower transformations are clustered using DBSCAN \cite{ester_density-based_1996}, and transformations outside the dominant cluster are considered outliers and removed.
From the remaining set, the best transformation is selected based on a metric combining fitness and inlier Root Mean Square Error (RMSE).
From this selected transformation, we iteratively apply scan matching for all frames where sufficient LiDAR overlap allows it, using the globally estimated tower position from the previous frame as the initial guess.
For frames without sufficient overlap or with unreliable results — such as those failing the clustering check or scoring poorly on the fitness/ inlier RMSE metric — the transformations are derived from KISS-ICP odometry and interpolated accordingly.
Even in the absence of direct sensor overlap, transformations can still be propagated under the assumption that the tower remains globally static.
\begin{figure}
    \centering
    \includegraphics[width=\linewidth]{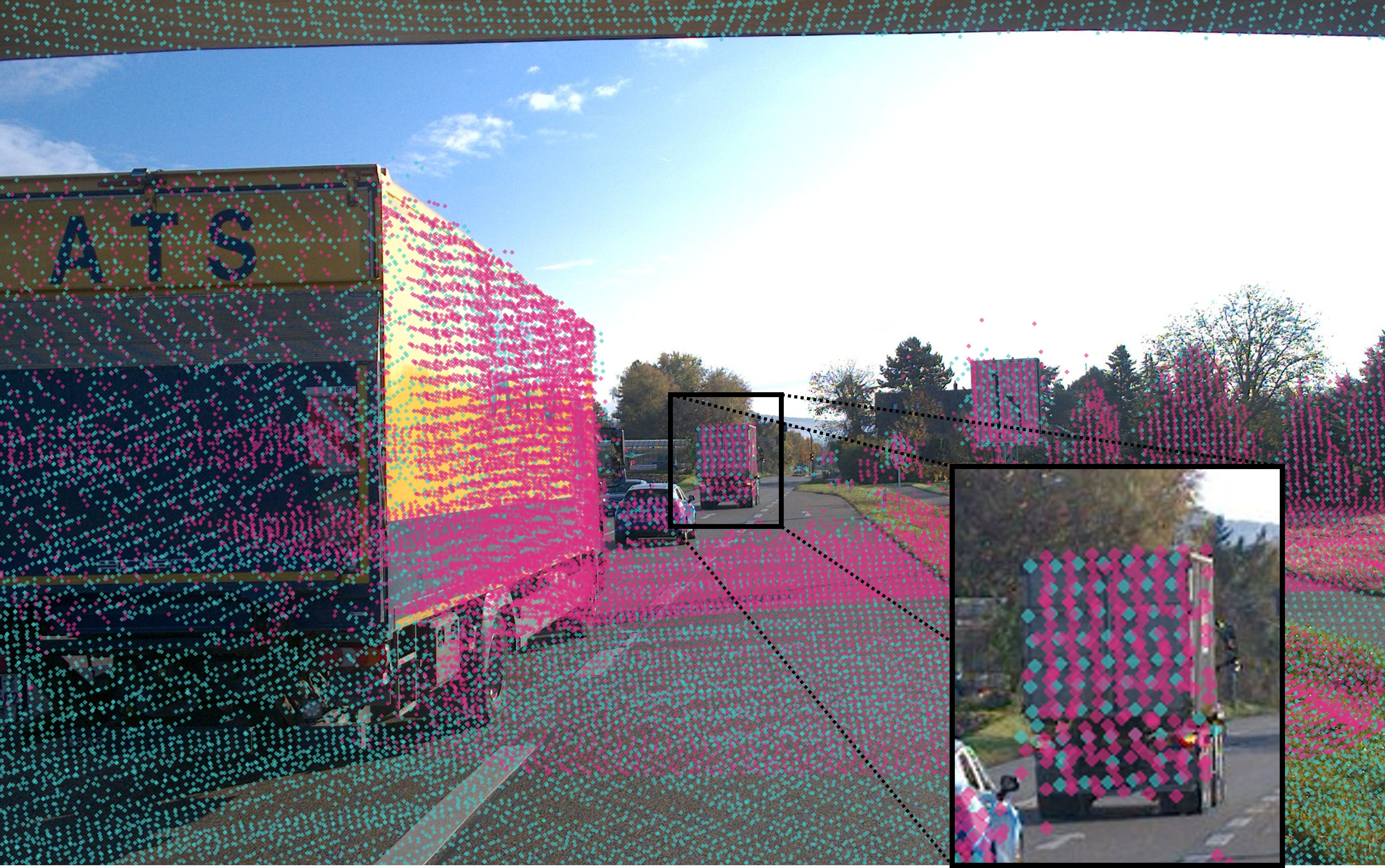}
    \caption{A rural road, captured during Scene \(6\), shows the vehicle’s front camera view at approximately \(45\) km/h. The turquoise points represent LiDAR measurements from the vehicle, while the pink points correspond to those from the observing infrastructure tower, located to the right of the vehicle. The vehicle's points are motion-compensated. The zoomed-in view of a distant truck illustrates the qualitative benchmark of the precise overlay of the data.}
    \label{fig:projection}
\end{figure}

Quantitatively evaluating cross-agent point cloud projections onto images is challenging. 
Therefore, we focused on qualitative assessments, such as the example shown in Figure \ref{fig:projection}.
At high speeds and with distant objects for both the vehicle and the infrastructure tower, the projection differences are expected to be most recognizable.
However, the qualitative results demonstrate a competitive performance, even under these challenging conditions.

\subsection{Anonymization.}
Data privacy is a critical consideration and a mandatory requirement in Europe.
To address this, we have anonymized our dataset, recognizing that this may influence the performance \cite{hukkelas_does_2023} of neural networks trained on it.
For anonymization, we developed a neural network called BlurScene capable of detecting faces and license plates (personal data) and applied a mosaic effect to these regions by default. 
The core architecture of the network is based on Faster R-CNN \cite{ren_faster_2015}, utilizing a ConvNeXt feature extractor \cite{liu_convnet_2022}. 
It achieves an inference time of \(0.2\) seconds per frame on a Nvidia GeForce RTX 4090.
The network was initially trained on the Wider Face \cite{yang_wider_2016} and Large-License-Plate-Detection \cite{elmenshawii_large-license-plate-detection-dataset_2023} datasets and fine-tuned using hand-annotated labels from our dataset.
Additionally, we created a validation set with \(40k\) labels, differentiating between identifiable and non-identifiable instances, to evaluate the recall specifically for identifiable labels.
This further allowed us to determine the optimal detection threshold (operating point) to minimize unnecessary masking.
We benchmarked our approach against EgoBlur \cite{raina_egoblur_2023} and achieved superior results on our dataset, successfully meeting the objective for our anonymization solution.
We achieved a recall of \(96\%\) on identifiable faces and \(98\%\) on readable license plates. The false positives of our detection have minimal impact on the images and lead to an unnecessary blur of approximately \(0.02\%\). Our solution outperforms EgoBlur, which achieved a recall of \(63\%\) for faces and \(71\%\) for license plates, when we allow for the same \(0.02\%\) unnecessary blur.
The BlurScene model, including weights and further analysis of the performance is available in the repository.
\section{Conclusion and Future Work}
We introduce CoopScenes, a multi-scene infrastructure-vehicle dataset aimed at fostering research in collective perception.
Notably, this is the first large-scale dataset of its kind to originate from Europe.
This dataset stands out with its unique features, including synchronized data across diverse scenarios, and is directly suitable for training neural networks on collective perception tasks.
This release includes a variety of scenes, particularly those related to public transportation, such as bus stations, pick-up/drop-off scenarios, and high-speed rural roads.
The dataset provides automatic annotation and anonymization pipelines, ensuring scalability for data extensions. 
All data and tools are publicly available. 
Additionally, we introduced agent transformations for spatial registration and a new level of temporal synchronization, setting a benchmark for time-sensitive applications.

Looking ahead, the dataset can be extended with additional scenes to enhance its diversity and with more annotations like 3D labels to broaden the task scope, while simultaneously improving the annotation pipelines. 
The anonymization pipeline introduced in this work will be explored in more detail in future research, including aspects such as model architecture, training strategy, and performance evaluation.
Furthermore, challenges such as real-time sensor registration and intention prediction for observing agents can be introduced and benchmarked to drive advancements in this domain.
We also plan to release further studies leveraging CoopScenes for cooperative perception tasks, such as Stixel-based object detection and sensor fusion. 


\begin{table*}[ht]
\centering
\caption{\textbf{Ego-Vehicle Sensor Specification} for Cameras including Stereo-Left (SL), Stereo-Right (SR), Back-Left (BL), Back-Right (BR), Front-Left (FL), Front-Right (FR), for LiDAR including TOP, Left (L), Right (R) and for INS.}
\begin{tabular}{l||C{2,8cm}C{2,8cm}|C{1,6cm}C{1,6cm}|C{2,3cm}}
 & \multicolumn{2}{c|}{Camera} & \multicolumn{2}{c|}{LiDAR} & INS \\
 & SL, SR, BL, BR  & FL, FR  & TOP  & L, R & GNSS/IMU \\
\hline\hline
Model & a2A1920-51gcPRO & a2A1920-51gcPRO & OS1 & OS0 & 3DM-GQ7 \\
Resolution & 1920 x 1200 & 1920 x 1200 & 1024 x 128 & 1024 x 128 & -\\
Frequency & 10Hz & 10Hz & 10Hz & 10Hz & 1kHz \\
HFoV/VFoV & 57.6\textdegree/37.7\textdegree & 79.1\textdegree/54.3\textdegree & 360\textdegree/45\textdegree & 360\textdegree/90\textdegree & -\\
Details & Exposure Time: 800$\mu s$, Focal Length: 6mm, Aperture: f/4.0 & Exposure Time: 800$\mu s$, Focal Length: 4mm, Aperture: f/4.0 & Range (10\%): 90m  & Range (10\%): 35m & Accuracy: 1cm RTK, 0,2\textdegree\ heading, 0,05\textdegree\ pitch/roll \\
\end{tabular}
\label{table:sensors_vehicle}
\end{table*}

\section*{Appendix}
This appendix provides additional technical details on the sensor setup, synchronization, calibration strategy, and development tools accompanying the CoopScenes dataset.

\subsection*{A. Setup Specification}
\textbf{Ego-vehicle.}
The sensor arrangement for the ego-vehicle is derived from a qualitative analysis focused on the informational needs within the vehicle cockpit, guided by human driving behaviors in public transportation.
This analysis resulted in a design featuring ten sensors, as detailed in table \ref{table:sensors_vehicle}, symmetrically aligned along the vehicle’s driving direction and approximately oriented horizontally.
Exceptions are the two LiDAR sensors at the front, one on the left and one on the right, which are tilted $45$\textdegree\ towards the respective sides.
Centrally positioned within the vehicle's sensor setup are the stereo camera and the top LiDAR sensor, with the entire arrangement referenced to the position of the top LiDAR.
The stereo camera, consisting of two aligned and flipped monocular cameras, faces the driving direction. 
It is mounted inside the bus, behind the top area of the windshield, an already proofed placement in various applications \cite{noauthor_distronic_nodate}.
To reduce reflections on the windshield, black molton fabric is placed on the dashboard below.
The LiDAR, located on the front section of the vehicle’s roof, provides a horizontal field of view (HFoV) of \(45\)\textdegree .
Our sensor setup does not adhere strictly to horizon alignment. 
Instead, it is engineered to ensure that the LiDAR coverage extends across the entire camera's field, including areas above the horizon.
However, given the design of our public transportation bus, we encounter limitations similar to those of an inclined sensor.
The vehicle's length and obstructions like the rear-mounted air conditioning system result in a constrained field of view for the sensor.

Compared to sedan cars, our bus's elevated setup height introduces close-range blind spots.
To mitigate this, we equipped the vehicle with two front LiDARs at the left and right corners, inclined both forwards and sideways by approximately \(45\)\textdegree . 
These LiDARs, with their 90-degree HFoV, provide a direct vertical view in front of the vehicle, effectively eliminating most shading issues.
To achieve a similar improvement in the camera system, we opted for a smaller \(4\) mm focal length, which enlarges the FoV at the expense of ranged detail.
For comprehensive lateral coverage, akin to the shoulder view, and to complete the near-complete surround view (excluding the rear), cameras with a \(6\) mm focal length were installed to function like modern digital side mirrors \cite{noauthor_digital_nodate}.

Navigation and motion sensing are realized by an INS, which combines GNSS and Inertial Measurement Unit (IMU) data using an adaptive Kalman filter. The system is positioned beneath the top-mounted LiDAR to optimize data integration and accuracy. To further improve the precision of the GNSS data, Real-Time Kinematic (RTK) positioning is employed, which utilizes RTCM correction messages transmitted from the SAPOS service.

\textbf{Infrastructure Tower}
The infrastructure tower is designed as a static counterpart to the moving ego-vehicle.
Its base consists of a modular foundation, featuring eight 25 kg weights on a metal structure, surmounted by a 3-meter pole.
Cross struts are installed to enhance the pole's stability. 
Designed for complete autonomy, the foundation includes compartments for car batteries, supplemented with a solar panel for power.
Furthermore, the tower is equipped with advanced communication technologies, including LTE/5G and V2X roadside unit (RSU), to support mobile connectivity.

Atop the pole, a metal plate hosts a central 360\textdegree\ LiDAR (horizontal) with GNSS antennas positioned on both sides, one for time synchronization and one for positioning.
Higher on the pole, we installed two camera-LiDAR units, each affixed to a freely movable ball joint, allowing for adjustable sensing directions tailored to each scene.
These cameras are identical to those used on the bus, each equipped with a 6 mm  lens. For the LiDAR component, we selected a solid-state device (SSD) to better align with the cameras' FoV.

In preparation for each scene recording, GNSS data are initially captured to accurately determine the position of the tower using a precise point positioning service \cite{noauthor_precise_2024}. More detailed information on the whole sensor setup is available in Table \ref{table:sensors_tower}.
\begin{table*}[ht]
\centering
\caption{\textbf{Infrastructure Tower Sensor Specification} for Cameras including View 1 (V1), View 2 (V2) for LiDAR including V1, V2, Upper Platform (UP) and for GNSS.}
\begin{tabular}{l||C{2,7cm}|C{2cm}C{1,6cm}|C{2cm}}
 & Camera & \multicolumn{2}{c|}{LiDAR} & GNSS \\
 & V1, V2 & V1, V2 & UP & GNSS\\
\hline\hline
Model & a2A1920-51gcPRO & Cube 1 Outdoor & OS2 & C099-F9P \\
Resolution & 1920 x 1200 & 400 x 51 & 1024 x 128 & -\\
Frequency & 10Hz & 10Hz & 10Hz & -\\
HFoV/VFoV & 57.6\textdegree/37.7\textdegree & 70\textdegree/30\textdegree & 360\textdegree/22,5\textdegree & -\\
Details & Exposure Time 800$\mu s$, Focal Length 6mm, Aperture f/4.0 &  Range (10\%): 30m & Range (10\%): 200m & PPP with \textgreater\ 1h data, accuracy variable \\
\end{tabular}
\label{table:sensors_tower}
\end{table*}

\subsection*{B. Synchronization Architecture}
The synchronization between the vehicle and infrastructure (tower) agents is carefully managed through both clock alignment and coordinated sensor triggering. 
We use the Precision Time Protocol (PTP) version 2 across our entire network to ensure high-precision time synchronization. 
The network architecture includes a dedicated PTP master clock, a PTP-capable switch configured as an end-to-end transparent clock, and sensors with built-in PTP support. 
To ensure consistency with global time, we rely on UTC timestamps derived from GPS signals.

The Inertial Navigation System (INS) on the vehicle maintains its own GPS synchronization independently. However, due to its high internal sampling rate of 1 kHz, no additional triggering mechanism is required for precise temporal alignment.

To ensure frame-level consistency across all sensors, triggering is coordinated around the vehicle’s TOP LiDAR. At each 0\textdegree\ rotation, this LiDAR generates a square-wave signal at $10$ Hz (period: $100$ ms, duty cycle: 50 \%), which is sent to a programmable logic controller (PLC). The PLC distributes the signal to all vehicle cameras to synchronize their image capture with the LiDAR rotation. The other LiDARs on the vehicle synchronize their own 0\textdegree\ timestamps via manufacturer software to align with the TOP LiDAR.

For the infrastructure agent, the $10$ Hz signal is regenerated based on the timestamps of the vehicle’s TOP LiDAR and used to trigger the tower cameras with microsecond-level precision. The two solid-state LiDARs on the tower do not support external triggering due to hardware limitations; however, they continuously record at maximum frequency, covering nearly the same field of view as the cameras. The 360-degree mechanical LiDAR on the tower supports either PTP synchronization or external triggering, but not both simultaneously. We opted for PTP to prioritize dataset-wide time alignment. Since each point in the LiDAR stream is individually timestamped, precise temporal information is retained even without external triggers.

\subsection*{C. Calibration Strategy}
The intrinsic calibration of the cameras was conducted using the multi-camera calibration functionality of the Kalibr toolbox \cite{maye_self-supervised_2013}. This self-calibrating algorithm surpasses other state-of-the-art methods in accuracy and provides a statistical quality report for each calibration session, which includes metrics like reprojection errors. The use of a robust calibration target, the \textit{aprilgrid} \cite{olson_apriltag_2011}, further enhances the precision of the calibration. Calibration was carried out under optimal weather conditions with a high-quality DIN A0 calibration target and an exposure time of 500 $\mu s$, selected to minimize motion blur and ensure the accuracy of the measurements. The recordings of the calibration process and the intrinsic results, including the stereo camera transformation and stereo-baseline, are available within the dataset.

For the extrinsic calibration of our setup, we utilized the velo2cam\_calibration suite \cite{beltran_automatic_2022}, which employs a method based on extracting reference points from a custom calibration target and determining their optimal rigid transformation through registration. We replicated this target using an aluminum composite panel, selected for its dimensional stability and precise reflectivity. Calibration was selectively conducted on the camera-LiDAR and LiDAR-LiDAR pairs necessary to define the geometry of each subsystem comprehensively, namely the vehicle and the infrastructure tower. For the tower, we conducted individual extrinsic calibrations for each scene to account for the flexible positioning of the camera-LiDAR pairs. 

To ensure geometric consistency, each subsystem is internally referenced to a designated root LiDAR. In the case of the vehicle, this reference is the TOP LiDAR, while for the tower, it is the UPPER\_PLATFORM LiDAR. All transformations within a subsystem are calculated relative to this root, enabling consistent spatial alignment and facilitating transformations between any two sensors. This relationship is expressed in Equation~\ref{form:trans}, where $T_{BA}$ denotes the transformation to the root sensor $A$ from any sensor $B$, and $T_{CA}$ from another sensor $C$.
\begin{align}
T_{BC} = T_{BA} \cdot T_{CA}^{-1}
\label{form:trans}
\end{align}

The transformation between the TOP LiDAR and the INS on the vehicle was precisely estimated through physical measurements, facilitated by the INS's placement directly below the TOP LiDAR. For the tower, the GNSS position was derived from CAD data. The transformation for each individual sensor to the root $T_{\mathcal{X}B}$ is included within the dataset.

\subsection*{D. Development Kit}
The primary objective of this development kit is to simplify data handling, enabling users to quickly download, install, and explore the dataset with minimal setup. 
Users benefit from straightforward, intuitive bindings, such as \texttt{frame[42].vehicle.lidars.TOP}, to directly access specific data entries like the TOP LiDAR point cloud. 
In addition, we provide several utility functions to address common processing needs and accelerate research workflows.

\textbf{Image functions.}  
By default, the dataset provides undistorted images to ensure consistency and usability for most vision-based tasks. 
If required, the original raw (distorted) images can also be accessed explicitly. 
This allows for custom undistortion or experiments with different camera parameters and models. 
Our default undistortion uses OpenCV \cite{noauthor_open_2024}, applying Lanczos interpolation over an 8×8 neighborhood. 
We also provide functionality for generating disparity or depth maps from stereo image pairs using the StereoSGBM algorithm from OpenCV.

\textbf{Sensor transformation.}  
To manage the various coordinate systems across sensors, we introduce a dedicated \texttt{Transformation} object that stores relevant information such as transformation matrix, translation, rotation (Euler angles), and the source and target frames. 
It supports standard operations like inversion and composition for flexible use across the sensor network.
Additionally, we provide a function specifically for transforming LiDAR points into a common coordinate frame—either relative to the local root of the corresponding agent or globally, relative to the tower UPPER\_PLATFORM LiDAR.

\textbf{LiDAR motion compensation.}  
To correct for motion-induced distortion in LiDAR scans, we apply a deskewing process using KISS-ICP \cite{vizzo_kiss-icp_2023}, leveraging the vehicle velocity estimated by the same pipeline. 
By default, all point clouds are returned in their undistorted form. 
The original raw scans are still available if explicitly requested, enabling deeper analysis or alternative motion compensation techniques.

\textbf{LiDAR-to-camera projection.}  
We provide tools to project 3D LiDAR point clouds onto the 2D image plane of a camera using rigid transformations. 
This creates a unified data representation that combines the high-resolution imagery of the camera with the spatial depth information from LiDAR. 
Our implementation includes mathematical transformation functions for alignment and calibration of the sensors, following established practices in the literature \cite{geiger_automatic_2012}.

\textbf{Hidden point removal.}  
Due to differing viewpoints and fields of view between LiDAR and cameras, projections may include points that are not actually visible in the image. 
To mitigate this, we implement a hidden point removal filter based on the Open3D implementation from \cite{katz_direct_2007}. 
An automatic parameter tuning mechanism is also included to adapt the filter to varying scene geometries and viewpoints.

\textbf{Visualization.}  
To further support data analysis, we provide a set of visualization tools—ranging from projecting multiple LiDAR point clouds in different colors onto the image plane, to generating interactive 3D visualizations with RGB-colored point clouds from all sensors. 
These tools are designed to be both easy to use and highly customizable, allowing for efficient exploration and debugging of multimodal data.



\addtolength{\textheight}{-3.0cm}   


\section*{Acknowledgment}
We gratefully acknowledge the Ministry of Transport of Baden-Württemberg for funding the AMEISE project, and Kaitos GmbH for their contributions to the development and implementation of the anonymization model.


\bibliographystyle{IEEEtran}
\bibliography{references,options} 

\end{document}